\documentclass[runningheads,orivec]{llncs}
\usepackage[T1]{fontenc}
\usepackage{newtxtext}
\usepackage{amsmath}
\usepackage[varvw]{newtxmath}
\usepackage{cite}
\usepackage{bm}

\usepackage{multirow}
\usepackage{graphicx}
\usepackage{textcomp}
\usepackage{xcolor}
\usepackage{multirow}
\usepackage{url}
\usepackage{booktabs}
\usepackage{wrapfig}
\usepackage{array}

\usepackage{colortbl}

\usepackage{makecell}
\usepackage{graphicx}
\usepackage{enumerate}
\usepackage{tabularx}
\usepackage{pifont}

\usepackage{hyperref}
\usepackage[caption=false]{subfig}
\newcommand{\subfigure}[2][]{\subfloat[#1]{#2}}
\usepackage{xurl}

\def\model{SciCREST}

\def\mydata{SciDAB}

\def\BibTeX{{\rm B\kern-.05em{\sc i\kern-.025em b}\kern-.08em
    T\kern-.1667em\lower.7ex\hbox{E}\kern-.125emX}}
\newsavebox{\tablefitbox}
\newcommand{\fitbox}[3]{\sbox{\tablefitbox}{#3}\ifdim\wd\tablefitbox>#1\relax
    \resizebox{#1}{#2}{\usebox{\tablefitbox}}\else
    \usebox{\tablefitbox}\fi}
\begin{document}

\title{\model: An Evidence-Constrained Reliable Agentic AI System for Scientific Dataset Analysis
}

\author{Qingqing Long \and Chenyang Zhao \and Xiaxun Xie \and Haotian Chen \and
Xiaolei Du \and Siyu Pu \and Xiaojie Zhu \and Xuezhi Wang  \and Hengshu Zhu \and Yuanchun Zhou}
\authorrunning{Q. Long et al.}
\titlerunning{SciCREST}
\institute{All authors are with the Computer Network
  Information Center, Chinese Academy of Sciences.}

\maketitle

\begin{abstract}
The fast-evolving field of Artificial Intelligence for Science (AI4S) has shown the critical role of scientific data in driving breakthroughs. This trend has led to various data-sharing services. As these platforms accumulate datasets, their role is moving beyond hosting and distribution. Researchers need platform-level analysis services to inspect files, understand basic file structure, and compute statistics. Existing data-sharing services and data-asking LLM agents usually treat files as flat inputs, making it hard to identify experimental-supported evidence, and thus produce unreliable results.
This paper introduces \model, an evidence-\underline{\textbf{c}}onstrained \underline{\textbf{r}}eliabl\underline{\textbf{e}} agentic AI \underline{\textbf{s}}ys\underline{\textbf{t}}em for scientific dataset analysis. \model~builds an experiment-organized evidence graph that captures file roles and relationships, and then selects query-specific evidence. A role-aware analytical executor derives the evidence execution constraints for reliable analytical steps.
\model~is deployed covering 10 million real-world datasets and 1 PB of source files. Offline and online evaluations show that \model~improves answer accuracy, evidence grounding, and hallucination control with production-level effectiveness. Our platform currently serves millions of users across more than 90 countries and is \href{https://ai.scidb.cn/en/page/adv1/new_session?recommend=true}{publicly accessible}.

\end{abstract}

\section{Introduction}
\label{Introduction}
\begingroup
\setlength{\columnsep}{8pt}
\setlength{\intextsep}{2pt}
\setlength{\abovecaptionskip}{3pt}
\begin{wrapfigure}{r}{.34\textwidth}
    \centering
    \includegraphics[width=.88\linewidth]{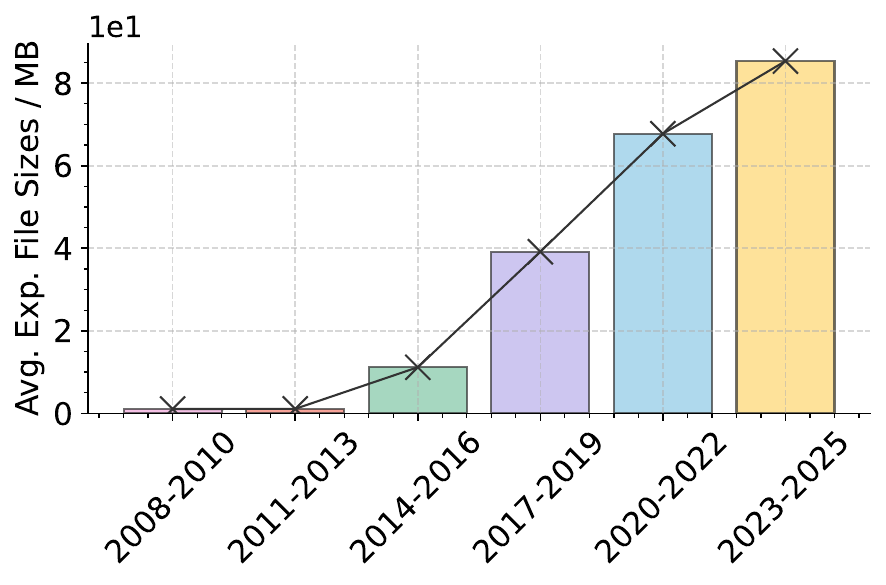}\par
    \vspace{-3mm}
    \caption{Yearly released scientific data file sizes on popular data sharing services, i.e., Zenodo~\cite{sicilia2017community}, ScienceDB~\cite{scienceDB1}, and DataCite~\cite{DataCite}.}
    \label{fig:intro_sizes}
\end{wrapfigure}
The fast-evolving field of Artificial Intelligence for Science (AI4S) has shown the critical role of scientific data in driving breakthroughs across various disciplines~\cite{ding2025scitoolagent}. Over the past decade, platforms such as Zenodo~\cite{sicilia2017community} and HuggingFace~\cite{jain2022hugging} have made scientific datasets easier to publish and access. The volume of newly released scientific data has increased substantially, as shown in Fig.~\ref{fig:intro_sizes}.

As data-sharing platforms continue to accumulate datasets, their role is moving beyond dataset distribution. Researchers who want to reuse a dataset~\cite{jain2022hugging} often need to inspect a file tree, identify relevant files, interpret metadata, compute statistics, and compare experimental conditions. These operations are difficult because scientific datasets are rarely released as a single table. They are often complex data packages containing raw observations, codebooks, protocols, derived statistics, etc~\cite{ai_scientist_lack_implementation}.
\par
\begin{wrapfigure}{R}{.42\textwidth}
    \centering
    \includegraphics[trim=3bp 8bp 3bp 1bp,clip,width=\linewidth]{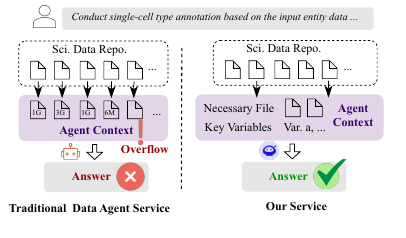}\par
    \vspace{-2.3mm}
    \caption{Our online service motivation.}
    \label{fig:intro_motivation}
\end{wrapfigure}
\noindent Platform-level services that provide quick and reliable dataset analysis, as illustrated in Fig.~\ref{fig:intro_motivation}, can effectively alleviate this burden~\cite{liew2016scientific}. They are therefore essential for turning shared datasets into practical support for scientific discovery.

LLM-based data-asking agents and  online services have already become a rapidly growing research topic for general data. 
These works can be generally categorized into three types:  \textit{code-driven agents}~\cite{qiao2023taskweaver}, \textit{tool-driven analyzers}~\cite{ding2025scitoolagent}, and \textit{workflow-driven agents}~\cite{AutoML_Agent}. Their detailed information is shown in Section 2. However, reliable dataset analysis introduces \textbf{unique challenges that are inadequately addressed}:
\par
\endgroup

(1) Scientific \textit{data packages} requires \textbf{experiment-organized evidence modeling}. As shown in Table \ref{tab:compare_product}, datasets in DataCite, Zenodo, and ScienceDB contain 2.25, 2.92, and 16.81 source files on average, respectively. These files are not independent but instead maintain strong referential and experimental connections through protocols, variables, and derived results, etc. 
However, existing works mainly focus on numerical computation or tool invocation. They treat files as a flat input set or retrieve them by textual similarity~\cite{ding2025scitoolagent}. They therefore miss role-aware supporting files, exceed the input context of LLMs, and produce hallucinated analysis when required evidence is absent or only implicitly connected~\cite{kramabench}.

(2) Scientific dataset analysis requires \textbf{intent-constrained evidence selection}. A scientific request often names visible entities, but correct analysis requires hidden supporting evidence. For example, a result table may require its schema, conditions, or upstream sources before its values can be correctly used. Existing code-driven and tool-driven agents mainly check whether code can run or whether a suitable tool is selected~\cite{ma2024sciagent}. They seldom infer the supporting evidence required by the requested operation. Scientific data agents, therefore, need role-aware selection over typed experimental relations, rather than a set of textually similar files.

(3) Reliable scientific analysis requires \textbf{operation and variable-level constraints}. Even if an agent selects a strict evidence set and successfully runs  code, the conclusion may still be unsupported~\cite{ai_scientist_lack_implementation}. An agent may perform unsupported joins, miss unit relations, or report a conclusion not supported by the executed data. Scientific data agents, therefore, need reliable operation-level evidence-constrained execution to final answers. Current online systems~\cite{jain2022hugging} do not consider such issues. Existing research works~\cite{hong2024metagpt,ding2025scitoolagent} also prioritize task success while ignoring reliable data asking.

In response, we develop \model, an Evidence-\textbf{C}onstrained \textbf{R}eliabl\textbf{e} Agentic AI \textbf{S}ys\textbf{t}em for scientific dataset repository analysis. \model~has been deployed on ScienceDB, one of the world's largest scientific dataset-sharing platforms~\cite{scienceDB1}, and \textbf{supports online analysis over 10 million real-world datasets}. For each request, \model~first grounds the query to a confirmed dataset and loads an offline-prebuilt evidence graph. The graph represents file roles and  experimental relations, so \model~can choose request-specific evidence and conditions instead of returning a flat file set. \model~then executes the analysis in a scoped code workspace where only admitted tables and metadata are exposed. 
For online efficiency, \model~uses precomputed summaries, projection-aware table loading, and chunk-based reading. 
\textbf{Our platform has been deployed online and currently serves millions of users across more than 90 countries.}
We summarize our contributions as follows:

\begin{itemize}
\vspace{-2mm}
    \item This paper designed and production-deployed an evidence-constrained reliable agentic service for scientific dataset analysis.
    \item The proposed framework has three key modules: an experiment-organized evidence graph constructor, an evidence selector, and a role-aware analytical executor.
    \item Offline and online experiments show that \model~improves answer accuracy, evidence grounding, hallucination control, and production-level effectiveness.
\end{itemize}

\vspace{-3mm}
\section{Related Work}
\vspace{-3.2mm}

\begingroup
\setlength{\columnsep}{8pt}
\setlength{\intextsep}{2pt}
\paragraph{\textbf{Scientific Dataset Sharing Platforms}}
\begin{wraptable}{r}{0.54\textwidth}
\centering
\setlength{\abovecaptionskip}{0pt}
\setlength{\belowcaptionskip}{3pt}
\caption{Comparison of popular data sharing services. CDA denotes the conversational data asking service. $*$ denotes that the result is estimated from the platform's top 1000 hot datasets.}
\label{tab:compare_product}
\footnotesize
\setlength{\tabcolsep}{1.2pt}
\fitbox{\linewidth}{!}{\begin{tabular}{cccccc}
\toprule
\textbf{Platform} & \textbf{Disciplines} & \textbf{Datasets} &
\makecell{\textbf{Source}\\\textbf{file types}} &
\makecell{\textbf{Avg. files/}\\\textbf{dataset}} & \textbf{CDA} \\
\midrule
    \makecell[l]{DataCite~\cite{ninkov2021datasets}} & $>$10 & 52,594,699 & $>$4 & $\sim$2.25$^*$ & \textcolor{red}{\ding{55}}\\  
    OpenAIRE~\cite{rettberg2012openaire} &  $>$10 & 8,382,956 & $>$5 & Unknown & \textcolor{red}{\ding{55}} \\
    Dataverse~\cite{magazine2011dataverse} &  $>$10 & $\sim$484K &  $>$12 & $\sim$8.11$^*$ & \textcolor{red}{\ding{55}}\\  
    CKAN~\cite{winn2013open} &  $>$10 & $>$484K & $>$10 & $\sim$2.26$^*$ & \textcolor{red}{\ding{55}}\\ 
    Dryad~\cite{he2016reuse} & $<$5 & $>$65K & $>$7 &  Unknown &   \textcolor{red}{\ding{55}}\\ 
    NCBI~\cite{geer2010ncbi} &  $<$3 & $\sim$1,000 &  $>$6 & Unknown & \textcolor{red}{\ding{55}}\\ 
    FigShare~\cite{thelwall2016figshare} & $<$3 & $\sim$380,000 & $>$6 &  Unknown  &  \textcolor{red}{\ding{55}}\\
    Zenodo~\cite{sicilia2017community} &  $<$5 &  6,598,220 & $>$380  & $\sim$2.92$^*$  & \textcolor{red}{\ding{55}}\\
    HuggingFace~\cite{jain2022hugging} & $<$3 & 461,199 & $>$9 & Unknown & \textcolor{green}{\ding{51}}  \\
    \midrule
    \textbf{\model~ (Ours)} & \textbf{All ($>$18)} & \textbf{10 Million} & $>$400 & 16.81 & \textcolor{green}{\ding{51}} \\ 
\bottomrule
\end{tabular}}
\end{wraptable}
Table~\ref{tab:compare_product} compares typical dataset-sharing services across five dimensions: (1) the number of supported disciplines, (2) the number of accessible datasets, (3) the number of source file types, (4) the average number of source files per dataset, and (5) the presence of an online Conversational Data Asking (CDA) service. Discipline coverage is estimated using the OpenAlex taxonomy~\cite{bordignon2024openalex}.
\model~covers more than 18 disciplines and supports an online CDA service.
\par
\endgroup

\vspace{-4mm}
\paragraph{\textbf{Code-Driven Agents and Harness}}
Code-driven agents~\cite{sun2025lambda,wang2024executable,qiao2023taskweaver} and harness~\cite{claude} generate executable programs for data analysis and refine them through execution feedback. These works turn reasoning steps into runnable code and often combine planning, interpretation, memory, and collaboration for data analysis. However, they usually view files as flat inputs. In contrast, \model~builds evidence graphs to select structurally supported files and reduce hallucinations.

\vspace{-4mm}
\paragraph{\textbf{Tool-driven Agents}}
Tool-driven agents~\cite{ding2025scitoolagent,ma2024sciagent}  emphasize identifying and invoking domain-specific tools to conduct data analysis tasks. Their core paradigm is to select and invoke domain-specific tools.
However, tool selection alone does not determine which files or conditions support a request. \model~performs evidence-constrained selection before execution.

\vspace{-4mm}
\paragraph{\textbf{Workflow-driven Agents}}
Workflow-driven agents automate data-science pipelines by decomposing tasks into stages~\cite{mlcopilot,hong2024metagpt}. This paradigm is effective when tasks have explicit objectives, such as classification and time-series prediction. Agents can select models and optimize predictive performance. However, scientific data asking is often open-ended rather than metric-driven~\cite{ma2024sciagent}.

\par\addvspace{12pt}
\noindent\begin{minipage}{\linewidth}
    \makeatletter\def\@captype{figure}\makeatother
    \vspace{-3.5mm}
    \centering
    \includegraphics[width=\linewidth,trim=0 8bp 0 0,clip]{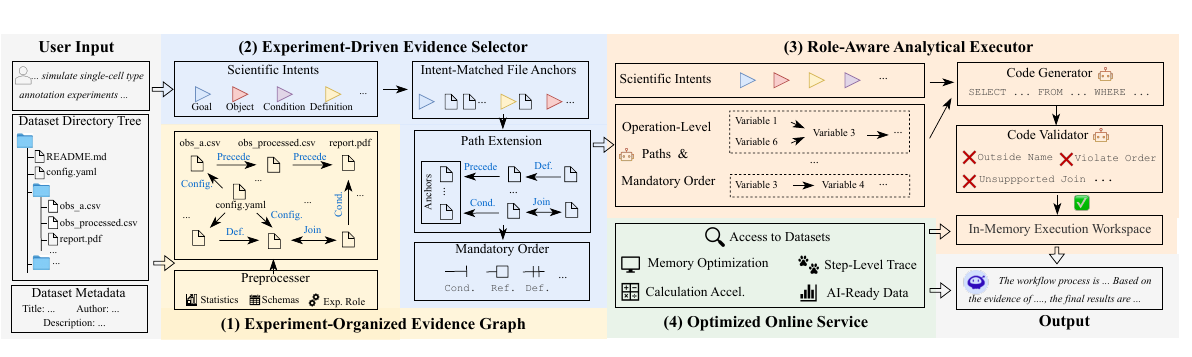}
    \par\setlength{\abovecaptionskip}{1pt}
    \caption{Technical framework of our \model~system.}
    \label{fig:framework}
\end{minipage}\par
\addvspace{15pt}

\section{Technical Details of \model}

The overall technical framework of \model~is shown in Fig.~\ref{fig:framework}. Examples from our public online system\href{https://ai.scidb.cn/en/page/adv1/new_session?recommend=true} are shown in Fig.~\ref{fig:online_system}.

\subsection{Experiment-Organized Evidence Graph}
\label{sec:model_kg}
A scientific dataset repository is usually organized around experiments, whose structural relationships differ from those in common graphs, such as paper citation or social networks. In practice, source files are naturally formed as a directory tree. However, directly giving LLMs all source files can bring irrelevant files while still missing crucial ones. It also consumes the LLM's limited context budget and makes reasoning slower and more difficult.

To address this issue, \model~builds an evidence graph over source file summaries, instead of raw file contents. Let $\mathcal{F}_d$ be the source files of dataset $d$. For each source file, preprocessing creates a source-summary node that records its file schema and path. We denote the summary-node set by $\mathcal{S}_d$. The evidence graph is represented as
$\mathcal{G}_d=(\mathcal{S}_d,\allowbreak{}\mathcal{R}_d,\allowbreak{}\mathbf{x}_d),$
where $\mathbf{x}_d$ contains dataset-level cues such as the title and description. The relation set $\mathcal{R}_d$ contains directed experimental relations. We represent a relation as $r=\langle s_i\rightarrow s_j,\tau,\kappa\rangle$, where $s_i\rightarrow s_j$ is a directed edge, $\tau$ is the relation type, and $\kappa$ is the confidence score from preprocessing. Inspired by scientific provenance studies~\cite{liew2016scientific}, file relations include the most common links. Specifically, documentation relations include \texttt{describe}, \texttt{summarize}, and \texttt{define}. Setup relations include \texttt{configure}, \texttt{use\_protocol}, and \texttt{condition}. Order relations include \texttt{derived\allowbreak{}\_\allowbreak{}from} and \texttt{precedes}. Alignment relations include \texttt{same\_\allowbreak{}sample}, \texttt{same\_\allowbreak{}condition},
and \texttt{joins\_\allowbreak{}on}. The relation-type vocabulary $\mathcal{T}$ is shared across datasets. The mapping $\mu_d:\mathcal{S}_d\rightarrow\mathcal{F}_d$ maps each summary node back to its original source file. It is then used in Section 3.2. 
\model~interprets each summary node by its experimental role $\rho(s)$, which describes how the corresponding file supports an analysis. A source file may act as an experimental protocol,  configuration, raw observation, etc.

\subsection{Experiment-Driven Evidence Selector}
\label{sec:claim_to_evidence_router}
Scientific data requests often verify a variable without naming the files that contain the observations or define the experimental conditions.  These requests are different from general data analysis tasks~\cite{dsbench}, as they always have a few certain files as input. 
Given a query $q_t$, an LLM parser extracts the analysis goal, typed entity mentions, and requested operation into a fixed-schema intent $\eta$. An LLM ranker $\phi$ scores the file summaries in $\mathcal{S}_d$ against this intent and the dataset context $\mathbf{x}_d$. The $K$ highest-ranked summaries form the anchor set $A = \operatorname{TopK}_{s\in\mathcal{S}_d} \phi(s;\eta,\mathbf{x}_d)$.

Then the selector determines which supporting roles are needed and expands anchors along relation paths. For example, a validation request may require the reported result, upstream source files, and the configuration or protocol that produced it. 
For each anchor, the selector uses its experimental role and the query intent to identify mandatory and optional evidence requirements. It admits a directed path from the anchor when the path matches the relation pattern, target role, and query entities specified by at least one requirement. Let $\mathcal{P}$ denote the admitted paths and $V(p)$ the nodes on path $p$. The anchors and all nodes on these paths form the evidence set $\mathcal{S}^{+} = A \cup \bigcup_{p\in\mathcal{P}} V(p)$. The selector maps these nodes to source files via $\mu_d$. It then passes the intent and selected evidence to the executor, together with mandatory requirements for each anchor and any operation dependencies generated by the selected paths.

\subsection{Role-Aware Analytical Executor}
Then we propose a role-aware analytical executor. It takes the chosen evidence subgraph as input, derives the constraints for reliable analysis, and finally carries out the  analytical steps. Unlike general tasks, scientific dataset repository analysis requires the executor to respect experimental roles and validity conditions, making reliable scientific execution more challenging.

The executor converts the selected files into logical relations $\mathcal{O}$ and derives the set of admissible joins $\mathcal{J}$ from the selected paths. An LLM planner with a fixed output schema translates the intent into typed operations such as column interpretation, normalization, and filtering. File schemas, experimental roles, conditions, and mandatory requirements constrain the plan. It uses the dependencies from the selector to group operations into stages. Each stage contains all pending operations whose prerequisites have been completed. If the dependencies form a cycle, \model~asks the user for clarification.
For each stage, an LLM generates a program $\pi$ in a restricted relational language. Before execution, an LLM validator checks that the relations read and the joins performed by $\pi$ satisfy $\operatorname{reads}(\pi)\subseteq\mathcal{O}$ and $\operatorname{joins}(\pi)\subseteq\mathcal{J}$. Codes are rejected before execution if they access relations outside the selected evidence, introduce unsupported joins, or violate the dependency order.

\subsection{Public Online System}
\paragraph{\textbf{Heterogeneous Source Files to AI-Ready Data}}
Scientific datasets often contain heterogeneous source files. Our platform~\footnote{https://www.scidb.cn/en} includes around 400 original file formats. To reduce online failures, we convert these files into AI-ready data. For practical coverage, we focus on the most common formats that account for the majority. Common structured formats include CSV, TSV, XLSX, HDF5, SAV, and Parquet. Common unstructured formats include PDF, DOCX, PPTX, JSON, and TXT.
Structured files are converted into standardized Parquet tables. Unstructured files are transformed into structured Markdown documents through an OCR parser~\cite{marker}.

\par\addvspace{6pt}
\noindent\begin{minipage}{\linewidth}
    \makeatletter\def\@captype{figure}\makeatother\vspace{-1mm}
    \centering
    \captionsetup[subfigure]{justification=centering}
    \subfigure[The Entrance]{\includegraphics[width=0.33\linewidth]{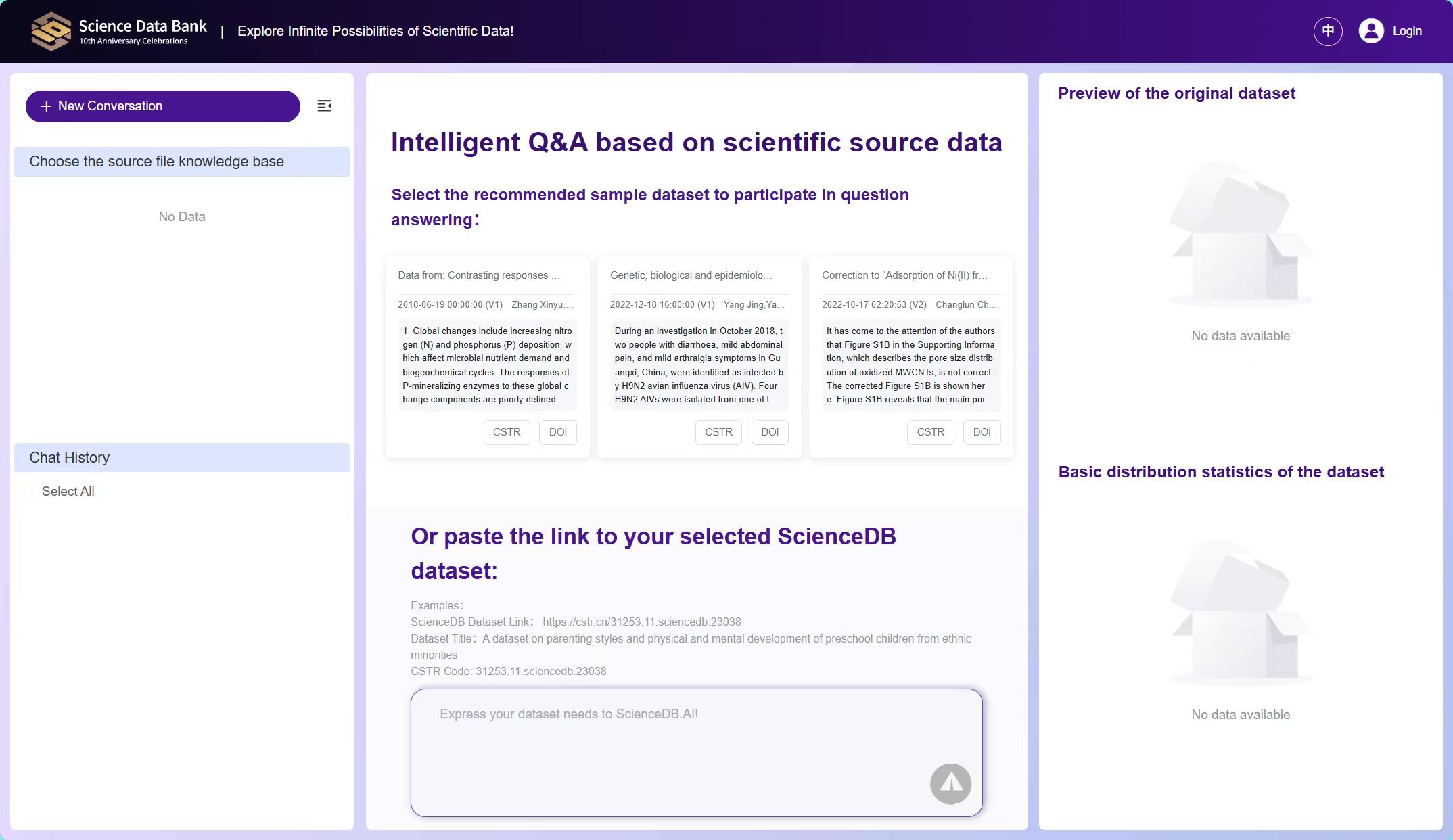}}\hspace{0.5mm}\subfigure[An online data analysis example]{\includegraphics[width=0.33\linewidth]{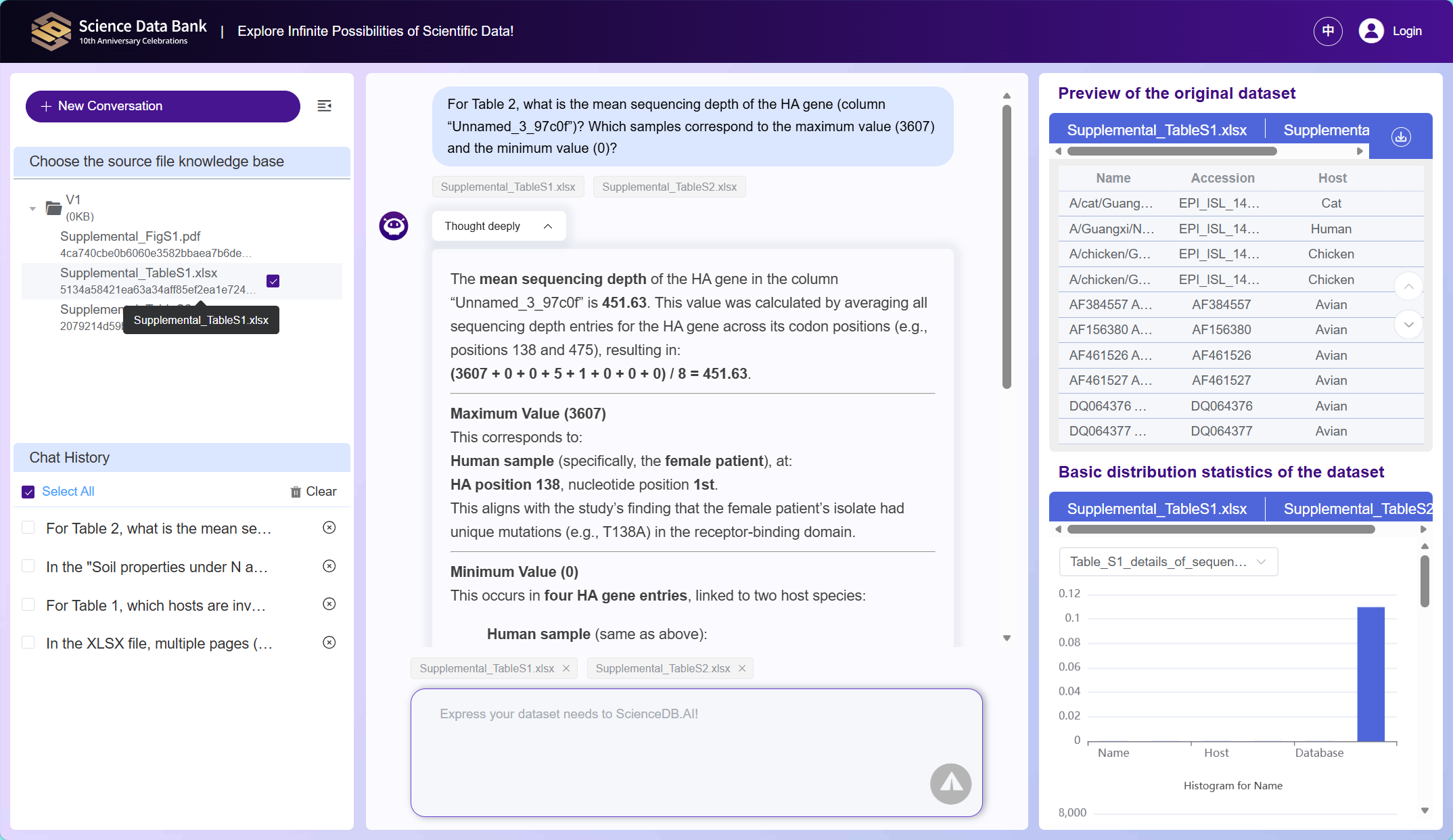}}\hspace{0.5mm}\subfigure[An online dataset example with source files]{\includegraphics[width=0.33\linewidth]{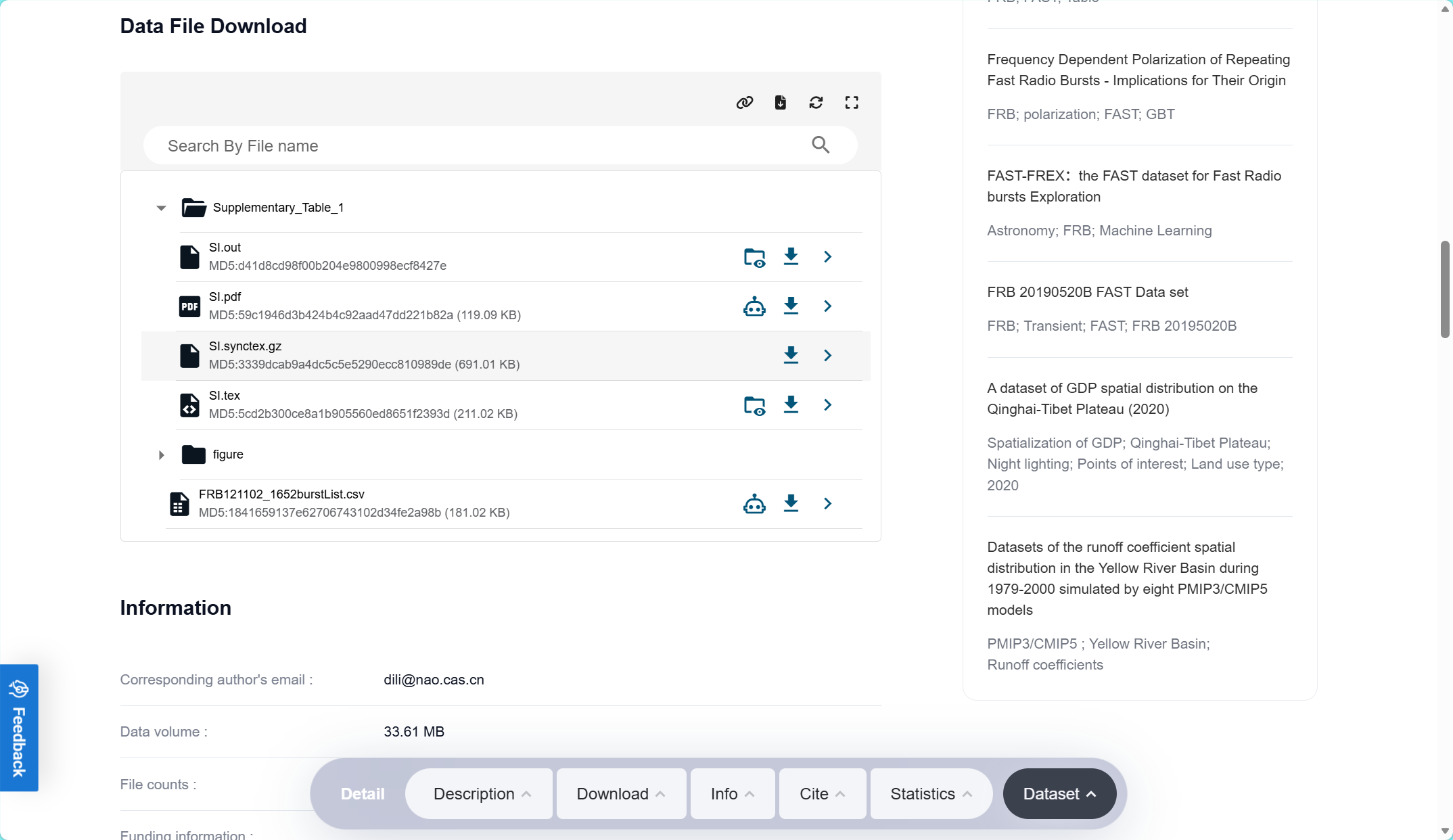}}
    \vspace{-2.5mm}
    \caption{Online examples of our \model. The platform is \href{https://ai.scidb.cn/en/page/adv1/new_session?recommend=true}{publicly accessible}.}
    \label{fig:online_system}
    \vspace{-3.5mm}
\end{minipage}\par
\addvspace{18pt}

\paragraph{\textbf{Access to Datasets}}
Users can begin analysis in two ways, as shown Fig.\ref{fig:online_system} (a). They may search for a dataset by name or description and choose a target dataset in an interactive way for later analysis. Alternatively, they may directly paste a dataset link to start. This dual entry supports both repository-scale discovery and link-based sharing in collaborative scientific research settings.

\vspace{-2mm}
\paragraph{\textbf{Acceleration Strategy for Online Calculation}}
\model~uses two strategies to reduce computation during online service. First, it adopts a summary-first policy. For each dataset, we precompute table schemas, basic statistics, document abstracts, and file descriptions. The agent checks these summaries before running costly tools. If the summaries provide enough evidence, it answers from the scoped context without loading the original files.
Second, table execution is projection-aware. The execution contract exposes the selected files, available schemas, and related table artifacts. The table worker can then load only the scoped files and query-relevant columns. This avoids unnecessary table loading and prevents the full source package from being used as the execution input.

\vspace{-2mm}
\paragraph{\textbf{Memory and Latency Optimization over Large Files}}
Large source files bring two challenges for online reasoning. Direct loading can cause memory overflow, and repeatedly loading and parsing have latency issues. In response, we build local analytical databases using DuckDB, which is lightweight and requires no service deployment. For each standardized Parquet table, we automatically create a DuckDB database. Table schemas are generated from file metadata, and data is inserted into the database. We then detect text columns and build indexes for them. These indexes support fuzzy queries during reasoning. The resulting DuckDB file serves as a fast retrieval layer for structured data and avoids repeated file parsing.
We address the memory issue with generator-based file reading in Pandas. Files are processed in chunks instead of being fully loaded into memory. Data is read only when needed and released after processing. This design prevents memory overflow and keeps memory usage stable during online reasoning.

\vspace{-2mm}
\paragraph{\textbf{Trace-Based Answer Verification}}
\model~records a step-level trace, which allows each user request to be traced back to executed queries, returned results, and admitted source files.
Then a verifier checks: (1) the answer must refer to the confirmed dataset, (2) cited files and variables must remain within the admitted scope, and (3) the answer must be supported by generated codes.

\section{Scientific Dataset Analysis Benchmark: \mydata}
\label{sec:dataset_bulid}
To provide a more comprehensive evaluation beyond existing public benchmarks~\cite{dsbench}, we construct the \textbf{Sci}entific \textbf{D}ataset \textbf{A}nalysis \textbf{B}enchmark (\textbf{\mydata}). 
The benchmark follows a four-level evaluation. Across all levels, hallucination probing tests whether an agent can recognize insufficient dataset evidence. 

\vspace{-2mm}
\paragraph{Level 1: \textbf{Evidence Structure Understanding}}
In scientific data exploration, researchers first screen candidate source files by examining organization, modalities, and experimental scope, as dataset-level metadata is often too coarse. This level evaluates whether an agent can form an evidence-aware overview of a dataset. Questions cover dataset composition, file types, modalities, directory structure, and scope. To provide domain background and references for prompting, we combine each dataset’s metadata with semantically similar questions retrieved from an MMLU~\cite{mmlu}-based vector database. MMLU is a question bank for college-level exams.

\vspace{-2mm}
\paragraph{Level 2: \textbf{Evidence Statistical Profiling}}
Before deeper analysis, researchers usually assess data availability and quality, which affect later conclusions. This level tests whether an agent can profile dataset contents while staying grounded in available evidence. Questions ask about data volume, completeness, missing values, and summary distributions. The question domain context is also derived from MMLU.

\vspace{-2mm}
\paragraph{Level 3: \textbf{Cross-Evidence Relational Analysis}}
Beyond basic statistics, researchers analyze quantitative relationships across variables, files, and experimental settings, such as variance under different configurations or performance across conditions. This level evaluates whether agents can reason over cross-evidence relations. We extend the Level 2 pipeline with metadata for column-pair relations and correlation indicators, producing questions that require multi-dimensional comparison and experimental interpretation rather than simple descriptive statistics.

\vspace{-2mm}
\paragraph{Level 4: \textbf{Experiment-Grounded Knowledge Discovery}}
After profiling and comparison, researchers derive scientific insights from information scattered across files, text documents, and variable definitions. This level focuses on experiment-grounded knowledge discovery, where answers must be supported by connected evidence rather than a single matched file. For each dataset $d$, we use the evidence graph $\mathcal{G}_d$ defined in Section~\ref{sec:model_kg}. The generator first identifies useful knowledge patterns as multi-hop paths $p:s_i\leadsto s_j$, then outputs the path sequence explicitly. A verifier checks structural validity and consistency with dataset content, filters invalid paths, and keeps only verified paths for multiple-choice question generation.

\vspace{-2mm}
\paragraph{Hallucination Probing at Each Level. \textbf{Know What is Unknown: Hallucination Probing}}
Hallucination is critical in scientific agents because unsupported answers can look plausible~\cite{ai_scientist_lack_implementation}, while existing evaluations often emphasize correctness over abstention~\cite{kramabench}. We therefore generate questions that are relevant to the dataset and domain but unanswerable from all provided resources, including descriptions, documentation, and data files. They are formatted as multiple-choice items with realistic, domain-consistent distractors. The correct option states that the answer is unavailable from the evidence. This probing is applied across Levels 1-4 rather than as a separate subset.

\section{Experiment}

\subsection{Experimental Settings}
\paragraph{\textbf{Dataset}}
The datasets used for questioning in \mydata~are randomly sampled from 10 million real-world datasets in ScienceDB. 
In \mydata, each of the level-specific evaluation subsets contains about 5,000 questions, yielding 25,846 questions in total. 
Each subset includes \textbf{both answerable and hallucination-probing questions}, and the latter are used to compute the Hallucination Rate (HR) within that subset.
Its detailed construction process is described in Section~\ref{sec:dataset_bulid}.      
For each task, \mydata~provides the dataset’s basic metadata and its  file tree as input. 

\paragraph{\textbf{Baselines}}
\setlength{\emergencystretch}{3em}
(1) \textit{Code-driven analyzers}, including CodeAct~\cite{wang2024executable} (ICML 2024), TaskWeaver~\cite{qiao2023taskweaver} (Arxiv 2023), LAMBDA~\cite{sun2025lambda} (JASA 2025), and DeepAnalyze~\cite{zhang2025deepanalyze} (PMLR 2026). 
(2) \textit{Tool-driven agents}, which focus on identifying and invoking domain-specific tools to answer user queries. We select  SciToolAgent~\cite{ding2025scitoolagent} (Nature Computational Science 2025) as the representative work.  (3) \textit{Workflow-driven agents}, 
including MLCopilot~\cite{mlcopilot} (ACL 2024) and MetaGPT~\cite{hong2024metagpt} (ICLR 2024).
(4) \textit{Harness}, including the Claude Code~\cite{claude} (2026) as a typical work.

\paragraph{\textbf{Implementation Details}}
We employ Qwen3.5~\cite{qwen3}  as the backend LLM of \model~and all baselines to deal with file-workflow planning, code generation, etc. The framework of \model~is implemented using Python and LangGraph. We record session states and step-level traces in PostgreSQL.
Both the dataset metadata (e.g., title, description) and the naturally structured file tree are provided as inputs to our model and all baselines. 
All baselines are evaluated under their default hyperparameters. 
We employ Qwen-Plus as the question generator to construct \mydata.

\subsection{Evaluation Protocols}
\label{sec:metrics_and_eval}
For answer quality, we report \textbf{Accuracy (Acc)} and \textbf{Reference Accuracy Rate (RAR)}. \textit{Acc} measures the correctness of the final answer, while \textit{RAR} measures whether the answer is well grounded in the provided references, considering evidence relevance, coverage, and logical use. 
For execution reliability and efficiency, we report \textbf{Success Rate (SR)}. A task is successful if the system returns a valid result without interruption. \textit{SR} is the proportion of completed tasks. 
For hallucination probing, we report \textbf{Hallucination Rate (HR)}. For each task level, HR is the percentage of hallucination-probing questions answered incorrectly at that level. A lower value indicates fewer unsupported responses.

\begingroup
\setlength{\intextsep}{8pt}
\begin{table}[!ht]
\centering
\caption{The overall performance comparison in \mydata. Accuracy (Acc), Hallucination Rate (HR), Reference Accuracy Rate (RAR), and Success Rate (SR) are evaluation metrics.}
\vspace{-2.5mm}
\label{tab:main_res1}
\setlength{\aboverulesep}{0.2ex}
\setlength{\belowrulesep}{0.3ex}
\resizebox{\linewidth}{!}{
\begin{tabular}{ccccccccccccccccc}
\toprule
\multirow{2}{*}{\textbf{Model}} & \multicolumn{4}{c}{\textbf{Structure Understanding}} & \multicolumn{4}{c}{\textbf{Statistical Profiling}} & \multicolumn{4}{c}{\textbf{Relational Analysis}} & \multicolumn{4}{c}{\textbf{Knowledge Discovery}} \\
\cmidrule(lr){2-5} \cmidrule(lr){6-9} \cmidrule(lr){10-13} \cmidrule(lr){14-17}
& \textbf{Acc} $\uparrow$ & \textbf{HR} $\downarrow$ & \textbf{RAR} $\uparrow$ & \textbf{SR} $\uparrow$ & \textbf{Acc} $\uparrow$ & \textbf{HR} $\downarrow$ & \textbf{RAR} $\uparrow$ & \textbf{SR} $\uparrow$ & \textbf{Acc} $\uparrow$ & \textbf{HR} $\downarrow$ & \textbf{RAR} $\uparrow$ & \textbf{SR} $\uparrow$ & \textbf{Acc} $\uparrow$ & \textbf{HR} $\downarrow$ & \textbf{RAR} $\uparrow$ & \textbf{SR} $\uparrow$ \\
\midrule
MLCopilot 
& 95.65 & 23.28 & 94.49 & \textbf{100.00} & 47.80 & 22.83 & 49.84 & 83.95 & 39.46 & 33.33 & 40.77 & 67.04 & 79.41 & 41.98 & 79.07 & 96.62\\
MetaGPT 
& 96.32 & 32.65 & 95.48 & \textbf{100.00} & 24.52 & 22.92 & 21.40 & 90.63 & 22.85 & 55.32 & 19.62 & 95.87 & 35.97 & 72.71 & 35.71 & \textbf{100.00}\\
\midrule
LAMBDA 
& 97.33 & 6.03 & 94.32 & \textbf{100.00} & 44.19 & 29.17 & 46.52 & 98.61 & 31.80 & 59.57 & 38.86 & 94.67 & 68.42 & 27.83 & 46.28 & \textbf{100.00}\\
CodeAct 
& 97.12 & 24.74 & 95.51 & \textbf{100.00} & 56.87 & 38.70 & 57.34 & 97.07 & 35.48 & 51.53 & 42.34 & 95.79 & 78.52 & 43.15 & 79.97 & 98.00\\
TaskWeaver 
& 89.94 & 25.79 & 90.59 & 99.29 & 35.70 & 33.47 & 33.30 & \textbf{100.00} & 24.93 & 54.78 & 23.33 & \textbf{100.00} & 69.46 & 72.89 & 68.99 & \textbf{100.00} \\
DeepAnalyze 
& 96.49 & 18.62 & 95.72 & \textbf{100.00} & 40.30 & 32.99 & 45.63 & \textbf{100.00} & 32.97 & 52.75 & 37.67 & \textbf{100.00} & 72.93 & 45.08 & 75.47 & \textbf{100.00}\\
\midrule
SciToolAgent 
& 88.35 & 5.56 & 89.97 & \textbf{100.00} & 29.10 & 10.26 & 32.54 & 48.32 & 25.43 & 36.36 & 32.24 & 38.67 & 72.98 & 50.99 & 72.46 & 38.41\\
\midrule
Harness
& 73.77 & 16.00 & 75.92 & \textbf{100.00}
& 19.20 & 16.00 & 23.00 & \textbf{100.00}
& 7.03 & 32.59 & 9.10 & \textbf{100.00}
& 78.13 & 39.20 & 78.97 & \textbf{100.00}\\
\midrule
\textbf{Ours} 
& \textbf{98.21} & \textbf{5.39} & \textbf{96.82} & \textbf{100.00} & \textbf{60.81} & \textbf{7.13} & \textbf{62.49} & \textbf{100.00} & \textbf{39.79} & \textbf{30.74} & \textbf{44.81} & \textbf{100.00} & \textbf{80.16} & \textbf{26.75} & \textbf{80.34} & \textbf{100.00}\\
\bottomrule
\end{tabular}
}
\end{table}
\endgroup

\subsection{Overall Performance}
Table~\ref{tab:main_res1} compares \model~with eight baselines on \mydata. Across all four task levels, \model~achieves the highest accuracy and reference accuracy rate (RAR), the lowest hallucination rate, and a 100\% success rate.
The largest gain of \model~in both accuracy and RAR is on Statistical Profiling. The results show the superiority of \model~in determining relationships among variables, files, and experimental conditions in a scientific dataset repository.

\newsavebox{\onlineABfigure}
\begin{lrbox}{\onlineABfigure}
    \begin{minipage}{.46\linewidth}
        \centering
        \includegraphics[width=\linewidth]{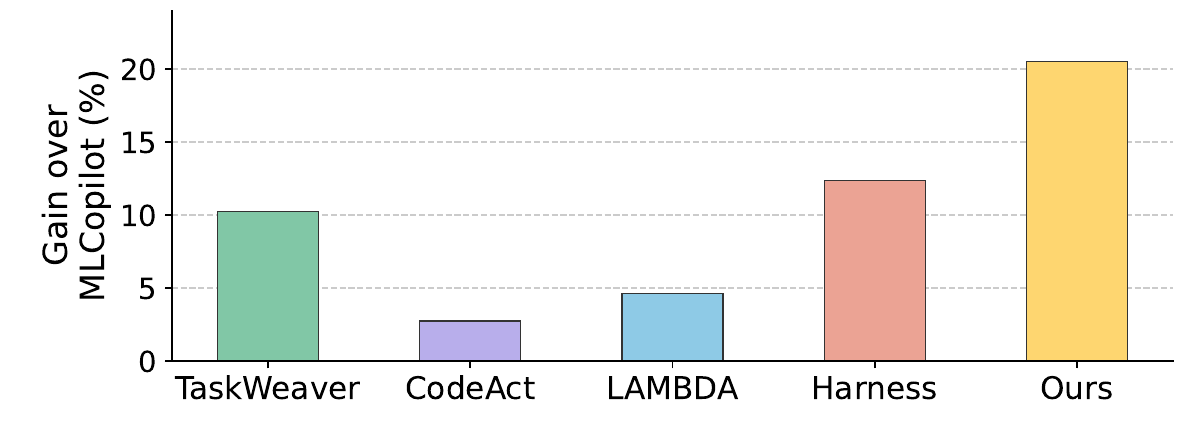}\par
        \setlength{\abovecaptionskip}{3pt}
        \makeatletter\def\@captype{figure}\makeatother
        \caption{Overall score gains over MLCopilot in the online A/B test.}
        \label{fig:online_res}
    \end{minipage}
\end{lrbox}

\subsection{Online A/B Test}
\begingroup
\setlength{\columnsep}{8pt}
\setlength{\intextsep}{2pt}
\begin{wrapfigure}{r}{\wd\onlineABfigure}
    \usebox{\onlineABfigure}
\end{wrapfigure}
We ran an online A/B test on ScienceDB. We report results for \model, TaskWeaver, CodeAct, and LAMBDA relative to MLCopilot. Task completion rate measures how often a system successfully answers a researcher's question. Average turns per task measures how many dialogue turns it takes to complete a task. Response latency per turn measures how long the system takes to respond. Question coverage is the proportion of question types the system supports, such as statistical computation, trend analysis, and causal inference.
An LLM judge evaluated the online session logs, and domain experts reviewed a subset to calibrate the judge's assessments. We combined the metrics into a weighted overall score. Results are shown in Fig.~\ref{fig:online_res}. \model~had the highest overall score.
\par
\endgroup

\section{Conclusion}
This paper introduces \model, an evidence-constrained agentic system for large-scale scientific dataset analysis. 
Offline and online results show that \model~produces more accurate, better-grounded answers with fewer hallucinations.

\bibliographystyle{splncs04}  
\bibliography{cite}

\end{document}